\documentclass[10pt,twocolumn,letterpaper]{article}

\usepackage{iccv}
\usepackage{times}
\usepackage{epsfig}
\usepackage{graphicx}
\usepackage{amsmath}
\usepackage{amssymb}
\usepackage{algorithm}
\usepackage{algorithmic}
\usepackage{multirow}
\usepackage{multicol}
\usepackage{booktabs}
\usepackage{subfig}
\usepackage{bbm}

\usepackage[pagebackref=true,breaklinks=true,letterpaper=true,colorlinks,bookmarks=false]{hyperref}
\iccvfinalcopy 

\def\iccvPaperID{****} 

\ificcvfinal\fi

\begin{document}

\title{Inter-class Discrepancy Alignment for Face Recognition}
\author
{Jiaheng Liu\thanks{means equal contribution} \textsuperscript{1}, Yudong Wu\footnotemark[1] \textsuperscript{1}, Yichao Wu\textsuperscript{1}, 
Zhenmao Li\textsuperscript{1},\\  Chen Ken\textsuperscript{1}, Ding Liang\textsuperscript{1}, Junjie Yan\textsuperscript{1}\\
\textsuperscript{1}SenseTime Group Limited \\
}
\maketitle
\ificcvfinal\thispagestyle{empty}\fi


\begin{abstract}

The field of face recognition (FR) has witnessed great progress with the surge of deep learning. 
Existing methods mainly focus on 
extracting discriminative features, and directly compute the cosine or L2 distance by the point-to-point way without considering the context information. 
In this study, we make a key observation
that the local context  represented by the similarities between the instance and  its inter-class neighbors~\footnote{In this paper, we refer inter-class neighbors to as support set.}
plays an important role for FR. 
Specifically, 
we attempt to incorporate the local information in the feature space into the metric,
and propose a unified framework called \textbf{Inter-class Discrepancy Alignment} (IDA), 
with two dedicated modules,
\textbf{Discrepancy Alignment Operator} (IDA-DAO) and \textbf{Support Set Estimation} (IDA-SSE).
IDA-DAO is used to 
align
the similarity scores considering the discrepancy between the images and 
its neighbors,
which is defined by adaptive 
support sets on the hypersphere. 
For practical inference, 
it is difficult to acquire 
support set during online inference. 
IDA-SSE can provide convincing 
inter-class neighbors
by 
introducing virtual candidate images generated with GAN.
Furthermore, 
we propose the 
learnable
IDA-SSE, 
which can implicitly give estimation without the need of any other images in the evaluation process.
The proposed IDA can be incorporated into existing FR systems seamlessly and efficiently. 
Extensive experiments demonstrate that this framework can 
1) significantly improve the accuracy, 
and 2) make the model robust to the face images of various distributions. 
Without bells and whistles, 
our method achieves state-of-the-art performance on multiple standard FR benchmarks.

\end{abstract}
\section{Introduction}
\begin{figure}[t]
\begin{center}
	\subfloat[The problem of conventional point-to-point (P2P) similarity.]
	{
		\label{fig:intro:a}
		\includegraphics[width=0.9\linewidth]{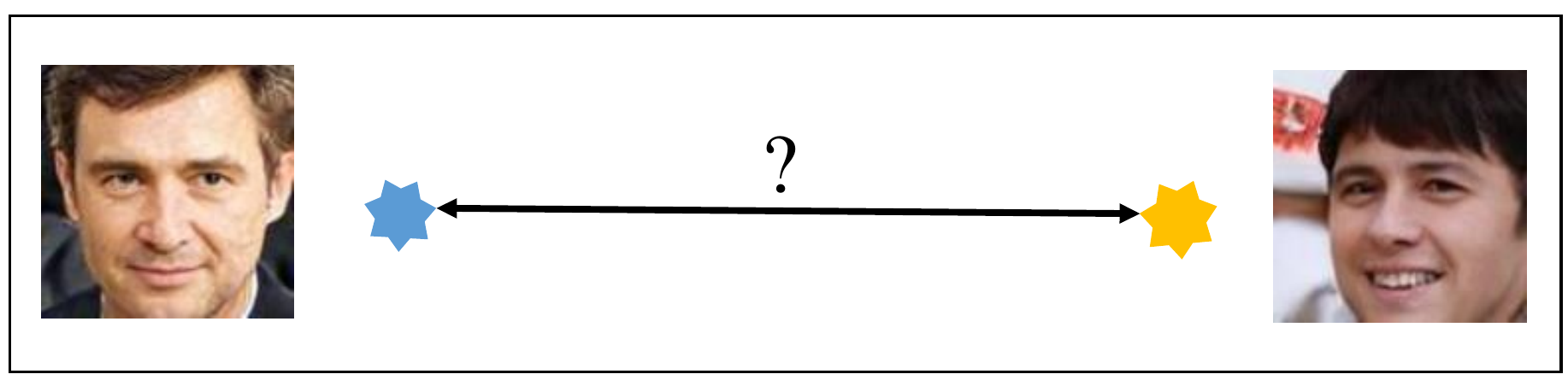}
	}
	\vspace{-10pt}
	\subfloat[Key observation of inter-class context information.]
	{
		\label{fig:intro:b}
		\includegraphics[width=0.9\linewidth]{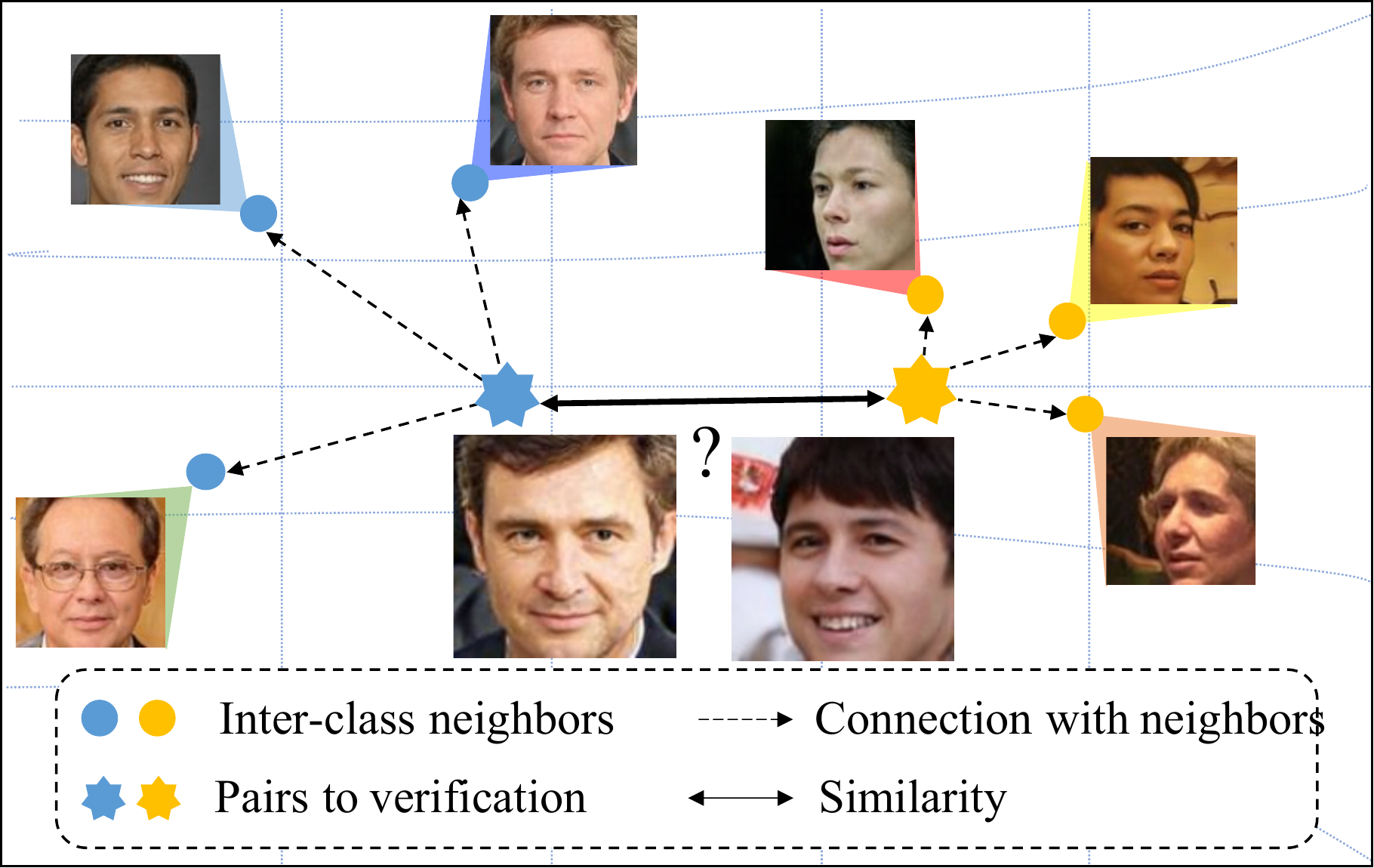}
	}
	\end{center}
	\vspace{-10pt}
\caption{
The motivation of our method. 
As shown in Fig.~\protect\subref{fig:intro:a}, 
the two face images are sampled from different identities. 
For P2P evaluation, 
we calculate the similarity scores only based on the isolated face images, 
and then determine the identity by 
thresholding cosine or L2 distance.
The conventional method may bring confusion, 
especially for face images of low quality or close relatives.
In Fig.~\protect\subref{fig:intro:b}.
we have a key observation:
the similarity 
scores
between an image and its inter-class neighbors
can offer
important cues to 
generate more convincing score. 
If we take into consideration that
the sample enclosed 
by the yellow star
inclines to be close to its neighbors of different identities, 
we could categorize this pair as an impostor one accurately.
}
\vspace{-10pt}
\label{fig:Intro}
\end{figure}
The problem of face recognition (FR) based on deep learning has been well investigated for decades~\cite{sun2014deep,sun2015deeply,wang1804deep}.
Much of this progress is  by 
large-scale training data~\cite{guo2016ms,yi2014learning,kemelmacher2016megaface}, 
deep neural network architectures~\cite{taigman2014deepface,he2016deep,hu2018squeeze}
and effective loss designs~\cite{ranjan2017l2,deng2019arcface,wang2018cosface,wang2018additive,zhang2019adacos,schroff2015facenet,chopra2005learning} 
for profiling the intrinsic property of FR. 
Despite the many efforts, 
these prior works often consider FR as a scenario of point-to-point (P2P) metric learning, 
where specific identities are determined by directly thresholding cosine or L2 distance between face image pairs.
However, 
for practical deployment,
we observe that proof of identity depends not only on the pair similarity, 
but also 
on the distribution around its neighbors 
on the hypersphere, 
where a simple P2P metric is insufficient to depict the complex local context for various samples.
%
There is a surge of interest to seek for incorporating the local information in the feature space into the metric.

A snapshot of a typical example is shown in Fig.~\ref{fig:intro:a},
where the similarity between isolated samples may cause confusions.
In conventional P2P evaluation process, 
the face verification
between face images from low quality or close relatives
could result in 
low-similarity for genuine pairs and 
high-similarity for impostor pairs.
It is usually insufficient to improve the capability of model itself, 
as the FR model alone cannot acquire the neighbor distribution during the online inference.
To solve this problem, 
as the core idea of this work, 
we have a key observation:
the similarity 
scores
between an image and its inter-class neighbors (referred to as support set) can offer
important cues to 
generate more convincing score. 
For example,
as shown in Fig.~\ref{fig:intro:b}, 
if we take into consideration that
the sample enclosed 
by the yellow star 
inclines to be close to its neighbors of different identities, 
we could categorize this pair as an impostor one,
so as to prevent from being \textit{false accept}. 




As far as we know, 
few approaches consider local context modelling in the field of FR.
Yu et al.~\cite{cheng2019graph} propose a feature aggregation method to 
utilize the mutual information between pairs in an offline FR testing environment. 
However, 
it is impossible to acquire the full information of the test set during online evaluation. 
PFE~\cite{shi2019probabilistic} can be seen as an alternative for context modelling, 
where only the intra-class information is considered. 


Motivated by the above analysis,
we propose a unified framework, 
named \textbf{Inter-class Discrepancy Alignment} (IDA), 
by leveraging the valuable inter-class information under the online test protocol.
IDA consists of two dedicated modules, including 
\textbf{Discrepancy Alignment Operator} (DAO) module and \textbf{Support Set Estimation} (SSE) module.
Specifically, 
for the Discrepancy Alignment Operator (DAO) module,
given a pair of face images, 
we adjust the similarity score between them using the context information (i.e., inter-class discrepancy) extracted from the support set.
For the Support Set Estimation (SSE) module,
two methods, the reference-based and learning-based methods,
are proposed to estimate the support set.
We employ GAN to produce a set of images from diverse identities referred as anchor image set.
In the reference-based method,
we search out the neighbors from the anchor image set in the feature space to generate the support set for each face image. 
%
In the learning-based method,
we leverage a learnable module 
to regress the inter-class discrepancy by implicitly constructing the support set. 
Extensive experiments on MegaFace~\cite{kemelmacher2016megaface}, LFW~\cite{huang2008labeled}, CALFW~\cite{zheng2017cross},  YTF~\cite{wolf2011face}, IJB-B~\cite{whitelam2017iarpa} and IJB-C~\cite{maze2018iarpa} show that 
our proposed IDA significantly improves the state-of-the-art performance of FR models. 
The proposed IDA has three appealing properties compared to the prior FR methods. 
\noindent \textbf{1) Generalizability.} Although many proposed methods attempt to improve the embedding quality~\cite{wang2018additive,deng2019arcface,zhang2019adacos}, 
the context modelling of feature space is rarely concerned. 
IDA makes use of support set to recalibrate the similarity scores for better generalization. 
\noindent \textbf{2) Latent.} For online evaluation, IDA does not need full access of the whole test sets explicitly compared with~\cite{cheng2019graph}. Instead, the context information can be estimated implicitly with the help of SSE.  
\noindent \textbf{3) Plug-and-play.} IDA is a simple and portable module, 
which can be readily integrated into any existing face recognition systems and boost their performance. 


In summary, our contributions could be summarized into three parts: 
\begin{itemize}
    \item [1)]  
    We investigate the effectiveness of inter-class local context in the filed of FR for the first time.  
    A unified framework called Inter-class Discrepancy Alignment (IDA) is proposed, 
    which consists of two modules, named as IDA-DAO and IDA-SSE respectively. 
    \item [2)]  
    The proposed Discrepancy Alignment Operator (DAO) is to recalibrate the similarities based on the local context information.
    Support Set Estimation (SSE) can provide convincing information of inter-class neighbors during online inference. 
    Our method can be readily integrated into existing FR systems.
    \item [3)]  
    Extensive experiments on several accessible benchmark datasets demonstrate the effectiveness of our proposed method for face recognition.
\end{itemize}
\section{Related works}
\noindent \textbf{Overview of Face Recognition.}
There are three essential factors for FR system, which are \textit{network architecture}~\cite{taigman2014deepface,6909640,sun2014deep,sun2015deeply,simonyan2014very,Szegedy_2015_CVPR,sun2015deepid3}, \textit{dataset}~\cite{guo2016ms,yi2014learning,kemelmacher2016megaface} and \textit{loss function}~\cite{schroff2015facenet,wen2016discriminative,zhang2017range,liu2016large,liu2017sphereface,wang2018cosface,wang2018additive,deng2019arcface,Peng_2019_ICCV,Jin_2019_ICCV,Wu_2020_CVPR}.
With the progress of CNN architectures, Deepface~\cite{taigman2014deepface} and DeepID~\cite{6909640} inherit typical object classification network for feature learning.
By combining both face identification and verification, DeepID2~\cite{sun2014deep} learns a robust and discriminative representation.
Furthermore, DeepID2+~\cite{sun2015deeply} increases the dimension of hidden representations and adds supervision to early convolutional layers.
Besides, VGGNet~\cite{simonyan2014very} and GoogleNet~\cite{Szegedy_2015_CVPR}, DeepID3~\cite{sun2015deepid3} further extend their work.
The promotion of FR also benefits from large scale of training data.
There are many authorized and widely-used datasets, including MS-Celeb-1M~\cite{guo2016ms}, CASIA-WebFace~\cite{yi2014learning}, MegaFace~\cite{kemelmacher2016megaface}, et al.

As for loss function, softmax is commonly used for image classification.
However, practical applications of FR are usually under the open-set protocol, where test categories are different from training categories~\cite{oh2016deep,sohn2016improved,wang2017deep}.
Most face recognition approaches utilize metric-targeted loss functions, such as triplet~\cite{schroff2015facenet}, which utilizes Euclidean distances to measure similarities between features.
Besides, center loss~\cite{wen2016discriminative}
and range loss~\cite{zhang2017range} are proposed to reduce intra-class variations via minimizing distances within each class.
Then angular-margin based loss functions are proposed to tackle the problem.
Specifically,
angular constraints are integrated into the softmax loss function to improve the learned face representation by L-softmax~\cite{liu2016large} and A-softmax~\cite{liu2017sphereface}.
CosFace~\cite{wang2018cosface}, AM-softmax~\cite{wang2018additive} and ArcFace~\cite{deng2019arcface},
which directly aim to maximize angular margins and employ a simpler and more intuitive way compared with aforementioned methods.
Despite the many efforts, 
these prior works often concentrate on improving the embedding quality, 
context knowledge from manifold distribution is neglected, 
which may convey rich information.

\noindent \textbf{Local Context in Face Clustering and Recognition.}
The local context information is significantly important for face clustering. 
\cite{rodriguez2014clustering} finds clusters by picking the samples with higher density than their neighbors. 
Similarly,
based on measuring density affinities between local
neighborhoods in the feature space, 
\cite{lin2018deep} proposes an unsupervised
clustering algorithm called Deep Density Clustering (DDC),
which obtains consistent performance improvement. 
\cite{frey2007clustering} uses affinity propagation to cluster images of faces with much lower error than other methods. 
\cite{sarfraz2019efficient} reveals that the first
neighbor of each sample is to discover large
chains and find the groups in the data.
In addition,
popular GCN-based face clustering methods \cite{yang2019learning,wang2019linkage,yang2020learning,ikami2018local} 
also take full advantage of the local context in the feature space around an instance,
which contains rich information about the linkage relationship between 
this instance and its neighbors. 
On the other hand, 
very few works have considered the context information for FR. 
By utilizing the mutual information between pairs
in the testing dataset, 
\cite{cheng2019graph} proposes an effective graph-based 
unsupervised feature aggregation method for FR. 
This method reveals the discriminative power of the local context, 
however, 
it is only applicable for offline evaluation. 
Cohort score normalization (CSN)~\cite{tistarelli2014use} has been used for face recognition based on traditional face descriptor by post-processing the raw matching score using the cohort samples and needs regression strategies to produce discriminative information.
PFE~\cite{shi2019probabilistic} can be seen as an alternative for context modeling, 
where only the intra-class information is considered. 
In this study, 
for the first time, 
we focus on leveraging the 
inter-class information under the online test protocol.

\section{Proposed Approach}
\subsection{Discrepancy Alignment Operator}
\label{sec:DAO}
In this subsection, we briefly illustrate the Discrepancy Alignment Operator (DAO) at first.
Subsequently, further analysis and instantiation of the operator are presented.

Commonly, FR approaches~\cite{chopra2005learning,schroff2015facenet,oh2016deep,deng2019arcface} learn an embedding per image using a CNN and directly adopt the P2P distance in the embedding space,
which corresponds to the face similarity during inference.
As mentioned above, we observe that cosine distance cannot accurately reflect the true similarity
and propose to incorporate the local inter-class discrepancy to calibrate the similarity score.
Specifically, we propose a new operator, which is named as \textit{Discrepancy Alignment Operator} (DAO), and formally:
\begin{equation}
\begin{split}
    \text{DAO}(\boldsymbol{f}_1, \boldsymbol{f}_2, \Psi_{\boldsymbol{f}_1}, \Psi_{\boldsymbol{f}_2}) =&\mathcal{G}(\langle\boldsymbol{f}_1, \boldsymbol{f}_2\rangle)\\&\cdot (\mathcal{F}(\boldsymbol{f}_1, \Psi_{\boldsymbol{f}_1}) + \mathcal{F}(\boldsymbol{f}_2, \Psi_{\boldsymbol{f}_2}))\ ,
\end{split}
\end{equation}
where, the $\mathcal{F}(\boldsymbol{f},\Psi_{\boldsymbol{f}})$ and $\mathcal{G}(\langle\boldsymbol{f}_1, \boldsymbol{f}_2\rangle)$ respectively represents the encapsulation of the local inter-class discrepancy and the inner product of two embeddings. The $\Psi_{\boldsymbol{f}} = (\boldsymbol{\hat{f}}_1, \boldsymbol{\hat{f}}_2,...,\boldsymbol{\hat{f}}_k)$, denotes the set of neighbors of the embedding in feature space, which we call ``support set''. The analysis and instantiation of the operator are presented in the following.

\noindent{\textbf{Inter-class discrepancy.}}
Here, we analyze the importance of inter-class discrepancy based on ArcFace~\cite{deng2019arcface}. 
It should be mentioned that we can acquire similar conclusions 
using other cosine-based softmax losses~\cite{liu2017sphereface,wang2018cosface}, triplet~\cite{schroff2015facenet} and contrastive~\cite{sun2015deeply} loss.
The ArcFace loss is as following:
\begin{equation}
    \mathcal{L}_{\text{ArcFace}} = - \log\frac{e^{z_{i,y_i}}}{\sum_{j=1}^{C}{e^{z_{i,j}}}},
\end{equation}
where $z_{i,j} = s \cdot \cos{(\theta_{i,j}+\mathbbm{1}\{j=y_i\} \cdot m)}$, and $\theta_{i,j}$ is the angle between j-th class weight and $\boldsymbol{f}_{i}$. The indicator function $\mathbbm{1}\{j=y_i\}$ returns 1 when $j = y_i$ and 0 otherwise. 
We can rewrite the ArcFace loss function as:

\begin{equation}
    \begin{aligned}
    \mathcal{L}_{\text{ArcFace}} = - \log\frac{1}{\sum_{j=1}^{C}{e^{(z_{i, j}-{z_{i,y_i}})}}} \ .
    \end{aligned}
\end{equation}

\begin{figure}[!htp]
    \begin{center}
    \subfloat[]{
	\label{fig:exp:a}
	\includegraphics[width=0.45\linewidth]{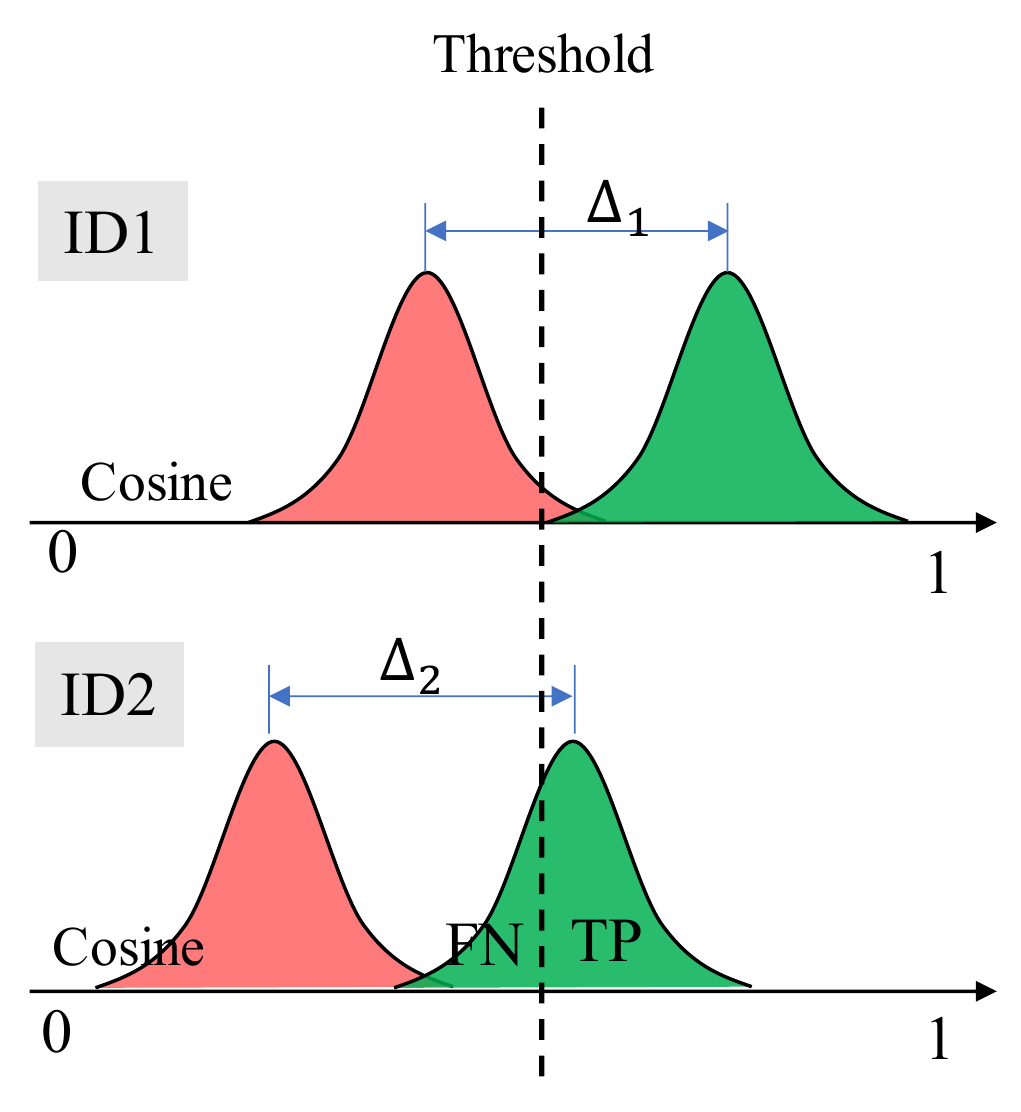}
	}
    \subfloat[]{
	\label{fig:exp:b}
	\includegraphics[width=0.45\linewidth]{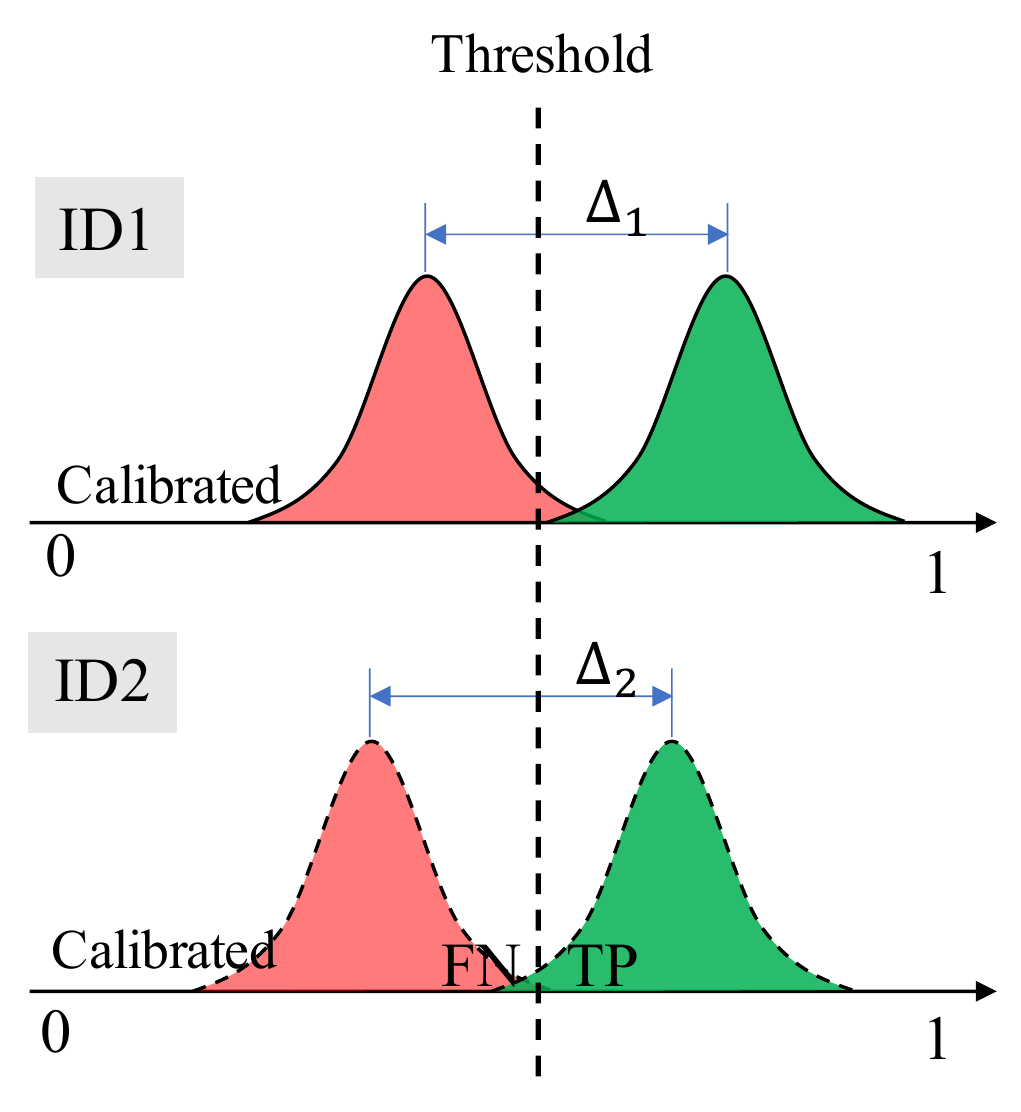}
	}
    \vspace{-10pt}
     \caption{Similarity distribution of pairs of different IDs. The green histogram represents the positive pairs, and the red represents the negatives. \protect\subref{fig:exp:a}: ID1 and ID2 have the same margin between positive and negative pairs, whereas the overall similarity of pairs of ID2 is less than ID1, which leads to  judging a large number of positive pairs as false negatives. \protect\subref{fig:exp:b}: When considering the local inter-class discrepancy, the scores of positive pairs of ID2 are calibrated to a high level.}
    \label{fig:exp}
    \vspace{-20pt}
    \end{center}
\end{figure}

Theoretically, the above loss functions are intrinsically relative-based, 
i.e., the gap between intra-class and inter-class similarity $z_{i,y_i} - z_{i, j}$ is proportional to the loss,
which can indicate how well the sample is optimized.
This optimization object may lead to a typical corner case, where samples converge to the same loss but correspond to the different cosine similarity.
For example, as shown in Fig.~\ref{fig:exp:a}, considering two IDs which have the same gap between positive pairs and negative pairs, whereas the overall similarity of pairs of ID2 is less than ID1. Therefore, a large number of positive pairs of ID2 are judged as false negative. During evaluation, the similarity between pair images are normally considered as $z_{i,y_i}$. To acquire more accurate scores, we need to introduce the item $z_{i, j}$, which represents the inter-class discrepancy, to the metric function. 
An additional statistical experiment is conducted in Sec.~\ref{sec:dis}.

\begin{figure}[!htp]
    \begin{center}
    \includegraphics[width=.85\linewidth]{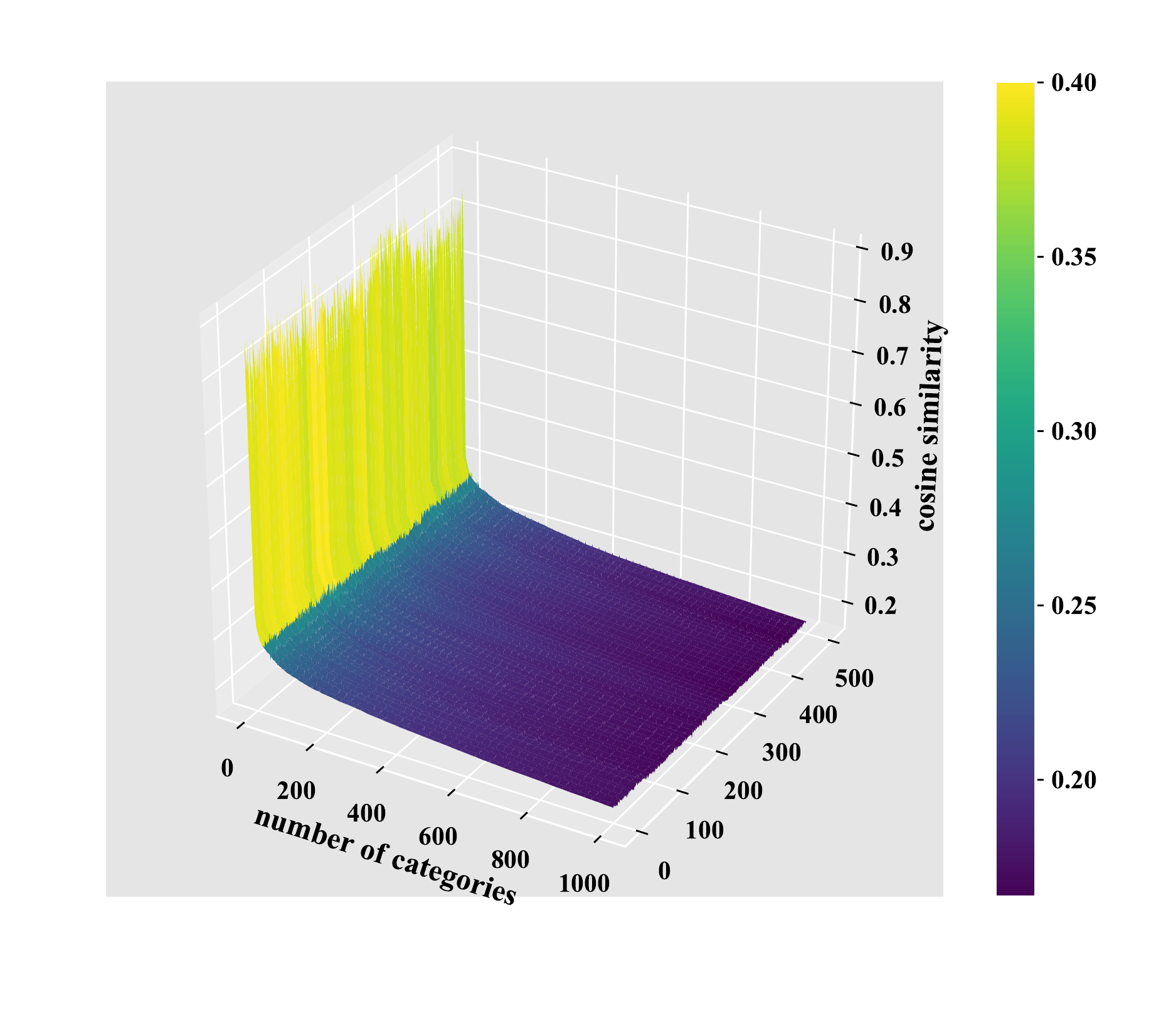}
    \vspace{-20pt}
     \caption{The cosine similarity distribution along category of each sample. We train the ResNet-50~\cite{he2016deep} on MS-Celeb-1M~\cite{guo2016ms} using ArcFace~\cite{deng2019arcface}, and randomly select 500 samples. Cosine similarity distribution of each sample along category is sketched, and each L-shaped curve represents a sample, i.e., $(\cos{\theta_{i,1}}, \cos{\theta_{i,2}},...,\cos{\theta_{i,C}})$.
     The $\cos{\theta_{i,j}}$ is the angle between i-th sample embedding and j-th class center. We tile all samples' curves along the y-axis. The similarities with other categories are descending sorted.}
     \label{fig:cosine_attenuation}
     \end{center}
     \vspace{-10pt}
\end{figure}

\noindent{\textbf{Support set.}}
In large-scale FR, the angles between the input sample and most other categories are around $\frac{\pi}{2}$, so their proportion of the inter-class discrepancy is minimal~\cite{zhang2018accelerated}. 
As shown in Fig~\ref{fig:cosine_attenuation}, the curve of inter-class cosine similarity of each sample quickly become flat and decay to $0$,
which indicates that the classes close to the input sample dominate the inter-class for this sample.
Hence, we factorize the inter-class discrepancy into $\sum_{j=1,j\neq y_i}^{k}{e^{z_{i,j}}}$ and $\sum_{j=k+1,j\neq y_i}^{n}{e^{z_{i,j}}}$ two parts. The first item consists of the closest $k$ inter-class neighbors of the sample in embedding space, i.e., $\Psi$. The $k$ is the size of the set. The similarities in the second term are approaching 0, so it can be simplified as $C - k - 1$, where $C$ is the number of classes. 

\noindent{\textbf{Instantiation of DAO.}}
Based on the above analysis, the $\text{DAO}(\boldsymbol{f}_1, \boldsymbol{f}_2, \Psi_{\boldsymbol{f}_1}, \Psi_{\boldsymbol{f}_2})$ can be intuitively instantiated as softmax-like:
\begin{equation}
\label{eq:dao}
    \text{DAO}(\boldsymbol{f}_1, \boldsymbol{f}_2, \Psi_{\boldsymbol{f}_1}, \Psi_{\boldsymbol{f}_2}) = e^{\tau \langle\boldsymbol{f}_1, \boldsymbol{f}_2\rangle} \cdot (\frac{1}{\sum_i^k{e^{\tau s_i^{1}}}} + \frac{1}{\sum_i^k{e ^{\tau s_i^{2}}}}),
\end{equation}
where $s_i^{1} = \langle\boldsymbol{f}_1, \boldsymbol{\hat{f}}_i^{1}\rangle \quad \text{for} \quad \boldsymbol{\hat{f}}_i^{1}\in \Psi_{\boldsymbol{f}_1}$, and $\tau$ is the hyperparameter. The formula satisfies the commutative property, i.e., $\text{DAO}(\boldsymbol{f}_1, \boldsymbol{f}_2, \Psi_{\boldsymbol{f}_1}, \Psi_{\boldsymbol{f}_2}) = \text{DAO}(\boldsymbol{f}_2, \boldsymbol{f}_1, \Psi_{\boldsymbol{f}_2}, \Psi_{\boldsymbol{f}_1})$. Here, the $\mathcal{F}(\boldsymbol{f},\Psi_{\boldsymbol{f}})$ is instantiated as $\frac{1}{\sum_i^k{e^{\tau s_i}}}$,
which indicates the density of IDs around the sample in embedding space. The denser the IDs distribute around the sample, the lower the obtained similarity is. This is reasonable that the density of the IDs reflects the discriminative power of this model.
Dense distribution means poor separability, so the similarity score should also be low.

\subsection{Support Set Estimation}
\label{sec:ref}
\begin{figure*}[t]
        \begin{center}
        \includegraphics[width=1.0\linewidth]{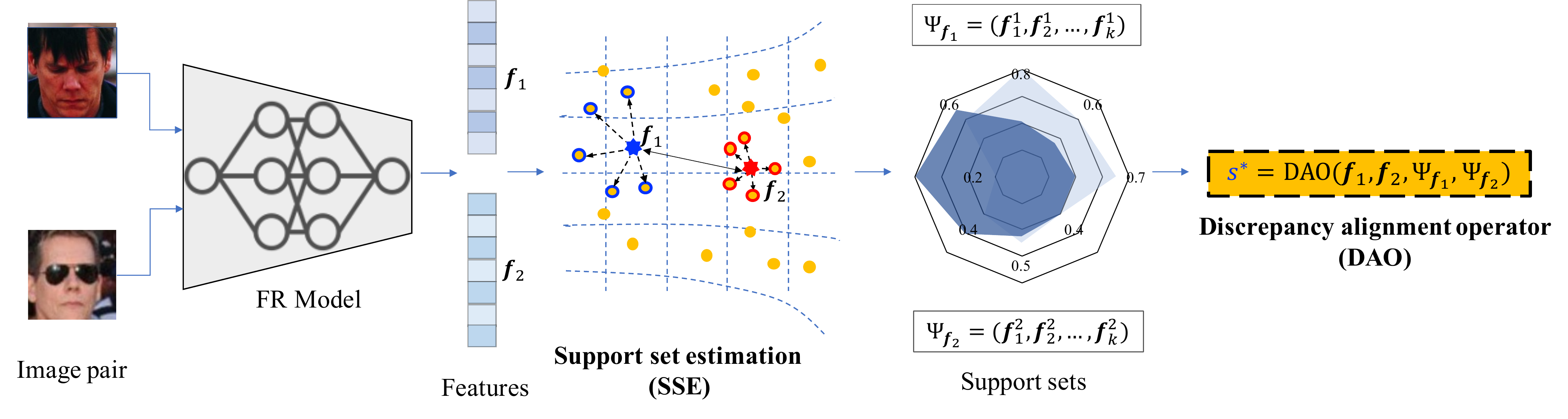}
        \vspace{-20pt}
         \caption{The complete verification process with reference-based support set estimation. Firstly, features of a face pair are extracted using the trained FR model. 
         Then we search for the support set of each feature in the anchor embedding set.
         The yellow points represent features in the anchor embedding set, and the blue and the red star denote the $\boldsymbol{f}_1$ and $\boldsymbol{f}_2$ from this face pair, respectively.
         Their nearest $k$ neighbors (i.e., support set) are shown by the dotted line. 
         Finally,
         $\text{DAO}$ takes the features and the support set from this pair as input,
         and return the calibrated score for face verification.
        }
        \label{big}
        
\vspace{-20pt}
        \end{center}
\end{figure*}
Accordingly, how to acquire the support set $\Psi_{\boldsymbol{f}}$ or the local density $\mathcal{F}(\boldsymbol{f},\Psi_{\boldsymbol{f}})$\footnote{As once the $\Psi$ is determined, the $\mathcal{F}(\boldsymbol{f},\Psi_{\boldsymbol{f}})$ can be calculated, so the $\Psi_{\boldsymbol{f}}$ and $\mathcal{F}(\boldsymbol{f},\Psi_{\boldsymbol{f}})$ can be used interchangeably.} of each sample during online inference is crucial.
In this subsection, we introduce two methods, the reference-based and the learning-based, to estimate the support set of sample.

\noindent{\textbf{Reference-based support set estimation.}} Acquiring support set is to estimate the local density $\mathcal{F}(\boldsymbol{f},\Psi_{\boldsymbol{f}})$ of the embedding distribution around the sample.
The entire embedding distribution of the FR model in feature space is hard to obtain.
We propose to generate a sampling from the embedding distribution.
Specifically, the FR model projects the face image to embedding:
\begin{equation}
    \boldsymbol{f} = \mathcal{M}_{\theta}(I), \ \ \ \text{for}\ \  I \in \mathcal{I}\ \  \text{and}\ \  \boldsymbol{f} \in \mathcal{E},
\end{equation}
where the $\mathcal{I}$ and $\mathcal{E}$ are the image space and embedding space.
The $\mathcal{M}_{\theta}$ is the FR model.
Hence, by uniformly sampling face images in $\mathcal{I}$ and then projecting to $\mathcal{E}$, we can obtain a sampling of the distribution.
We call the sampled image set the ``anchor image set'', denoted as $\mathcal{A}_I$.
The corresponding embedding set is named as ``anchor embedding set'', denoted as $\mathcal{A}_E$.
When verifying a face pair, corresponding support sets of the two features are obtained by searching the $\mathcal{A}_E$,
and the cosine similarity is calibrated through the DAO. The algorithm can be concluded in Algorithm~\ref{alg:Framwork}.

\begin{algorithm}[h] 
\caption{Reference-based support set estimation.} 
\label{alg:Framwork} 
\begin{algorithmic}[1] 
\REQUIRE ~~\\ 
two face images $I_1$ and $I_2$;\\
the trained FR model $\mathcal{M}_{\theta}$;
the anchor embedding set $\mathcal{A}_E$
\ENSURE ~~\\ 
normalized similarity of the two faces $s^*$;
\STATE Extracting features of two images, i.e., $\boldsymbol{f}_1 = \mathcal{M}_{\theta}(I_1)$, $\boldsymbol{f}_2 = \mathcal{M}_{\theta}(I_2)$; 
\STATE Acquiring the support set $\Psi_{\boldsymbol{f}_1}$ by searching for the top-k maximum similarity scores of $\boldsymbol{f}_1$ in the $\mathcal{A}_E$; 
\STATE Acquiring the support set $\Psi_{\boldsymbol{f}_2}$;
\RETURN $s^* = \text{DAO}(\boldsymbol{f}_1, \boldsymbol{f}_2, \Psi_{\boldsymbol{f}_1}, \Psi_{\boldsymbol{f}_2})$; 
\end{algorithmic}
\end{algorithm}
It is difficult to achieve uniform sampling,
and the function of the anchor image set is to obtain the embedding distribution of the FR model.
In practice, a straight way to construct the anchor image set is randomly sampling from the training set.
We call the resulting anchor feature set as the ``real-db''. Additionally, we propose to use fake face images which generated by GAN~\cite{goodfellow2014generative,karras2019style} to construct the set. We name this as the ``fake-db''. Compared with the real-db, fake images do not contain privacy information, which is more conducive to practical use and dissemination. Besides, the IDs of the fake-db will not conflict with the test sample.
In the following experiment section, the fake-db is demonstrated to have comparable performance with the real-db.
Furthermore,
the effectiveness of fake-db also shows that the specific ID is not important for estimating support set.

Compared with the original comparison process, the additional computational overhead is the process of searching for the support set and the DAO.
For the process of searching, the size of anchor embedding set is small (usually no more than 100000),
and off-the-shelf nearest neighbor search libraries can be used to reduce the time consumption~\cite{malkov2018efficient,JDH17}.
The complete verification process is also sketched in Fig.~\ref{big}.

\noindent{\textbf{Learning-based support set estimation.}}
\label{sec:learn}
In addition to obtaining the support set by querying the anchor embedding set during each inference,
we also innovatively proposed a learning-based method, in which the $\mathcal{F}(\boldsymbol{f},\Psi_{\boldsymbol{f}})$ is directly predicted by the FR model.
Specifically, an additional support set regression (SSR) module is adopted to learn the $\mathcal{F}(\Psi_{\boldsymbol{f}})$.
Formally the loss function to minimize is:
\begin{equation}
    \mathbb{L} = {\left\|M_{\text{SSR}}(\boldsymbol{f}) - \mathcal{F}(\boldsymbol{f},\Psi_{\boldsymbol{f}}) \right\|_{2}^{2}}\ .
\end{equation}
The SSR module consists of two fully connected layers,
so the increase in computational overhead is negligible. It takes face embedding as input and predicts the estimated $\mathcal{F}(\boldsymbol{f},\Psi_{\boldsymbol{f}})$.
Inspired by~\cite{shi2019probabilistic}, stage-wise training strategy is adopted: given a pre-trained FR model, we fix its parameters, and only optimize the SSR module to learn the $\mathcal{F}(\boldsymbol{f},\Psi_{\boldsymbol{f}})$. This strategy is conducive to rapid experimentation.
Besides, when the SSR module is trained on the same dataset of the FR model, this stage-wise training strategy provides a more fair comparison between the proposed method and the original model.

\subsection{Discussions}
\label{sec:dis}
\begin{figure}[htbp]
\noindent
\begin{minipage}{.35\linewidth}
\centering
\subfloat[training set]{\label{fig:domain:a}\includegraphics[width=0.98\linewidth]{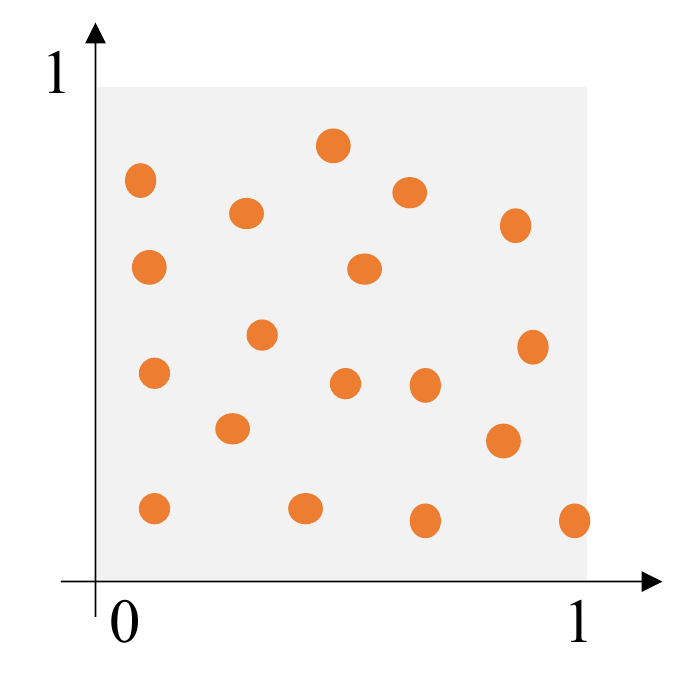}}
\end{minipage}%
\begin{minipage}{.35\linewidth}
\centering
\subfloat[testing set]{\label{fig:domain:b}\includegraphics[width=.98\linewidth]{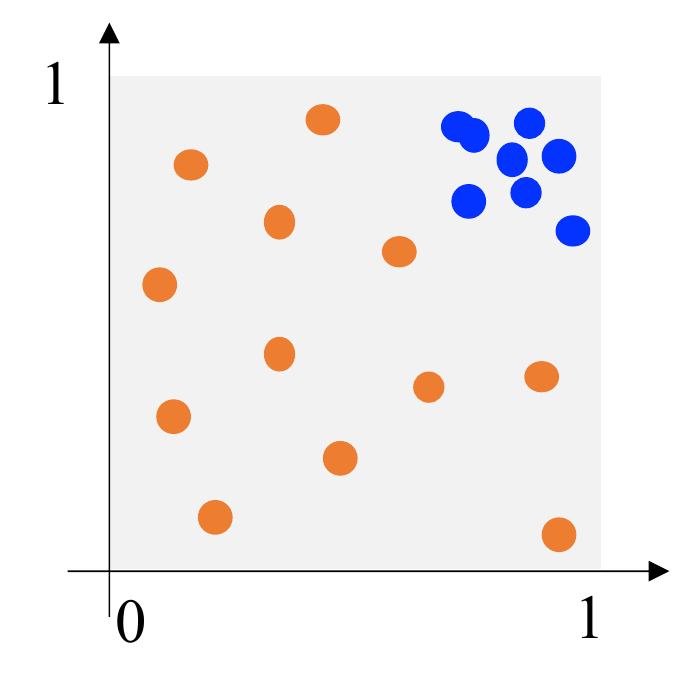}}
\end{minipage}
\centering
\vspace{-5pt}
\caption{Distribution of ID center in feature space. We train the FR model only using the yellow race faces. Each point represents a ID center, the yellow refers to the yellow races and the blue refers to the black. \protect\subref{fig:domain:a}: the distribution of ID centers on training data. \protect\subref{fig:domain:b}: the distribution ID centers of data that not used in training.}
\label{fig:domain}
\vspace{-3mm}
\end{figure}

\noindent \textbf{What causes the diverse local inter-class discrepancy?} 
 Different local inter-class discrepancy means that in feature space, the distribution of face embeddings on the hypersphere is not equally spaced. We believe that the main reason for this is the non-uniform training data, which causes the FR model to be biased towards different domain. For example, the model only trained on the yellow race faces will distribute the yellow race IDs scattered and evenly in the feature space. Whereas, the black race faces are embedded densely and ambiguously, just like a large cluster, as shown in Fig~\ref{fig:domain}. Here, We exemplified the bias of the model by extremely using different races as different domains. Actually, the domain of data is significantly diverse especially in unconstrained scenario, and the boundary is vague, difficult to define, even the domain itself cannot be considered discrete or enumerated. Consequently, it is tricky to reduce this misalignment from the data perspective.

\noindent \textbf{Why plug-and-play?} 
It is convenient to incorporate IDA into existing frameworks.  
Firstly, 
the proposed IDA does not change the way of feature extraction, 
thus, 
IDA can be combined with any modern network architectures, 
such as VGGNet~\cite{simonyan2014very}, GoogleNet~\cite{Szegedy_2015_CVPR} and ResNet~\cite{he2016deep}.
Secondly, 
IDA does not change the feature distribution on the hypersphere. 
Most FR loss functions~\cite{liu2017sphereface,wang2018cosface,deng2019arcface,wang2018support,zhao2019regularface}aims to 
learn discriminative and deterministic representations.
For these point-wise methods, 
our method is easy to refine accuracy with an extra inter-class discrepancy alignment in the testing stage.
Recently, 
some approaches have been proposed to utilize the information of feature distribution~\cite{shi2019probabilistic,chang2020data,ali2020biometricnet} for further improvement. 
Take PFE~\cite{shi2019probabilistic} for an example, 
the proposed method aims to model the intra-class variance,
while IDA focus on the inter-class context modelling.
Actually, 
our approach is promising to be combined with the PFE to incorporate both the inter-class and intra-class information during inference, 
which we will investigate in the experiment.

\noindent\textbf{Discussion with cohort score normalization.}
Cohort score normalization (CSN)~\cite{tistarelli2014use} has been used for face recognition by post-processing the raw matching score using the cohort samples.
Although CSN shares some common 
ideas with the proposed IDA, 
our contributions are still sufficient:
(1) Our method is based on the SOTA framework (consisting of modern architectures and losses of high performance), 
where the extracted features are distributed on the specified hypersphere. 
From the perspective of feature distribution as shown in Fig.~\ref{fig:cosine_attenuation}, 
we have key observations about the effectiveness of inter-class discrepancy for FR.
While for most CSN based methods, 
the system is usually built upon the traditional facial descriptor. 
Although the processes are similar, 
the motivation is quite different, and no discussion about the feature distribution is supplied in CSN. 
(2) CSN tries to exploit the patterns from sorted cohort scores and needs regression strategies to produce discriminative information, 
while 
our method is plug-and-play 
following Eq.~\ref{eq:dao} without external regression process.
Moreover, 
we propose a novel learning-based support set estimation module, 
which can abandon the anchor embedding set (cohort set in CSN) during testing.
(3) We have conducted comprehensive experiments and 
achieved significant performance improvements on multiple large-scale benchmark datasets.

\section{Experiments}
In this section, we conduct extensive experiments to demonstrate the effectiveness of our proposed method.
Then, we conduct a detailed ablation study to further analyse our method.
\subsection{Implementation details}
\textbf{Datasets.}
For the training datasets,
we employ the CASIA-WebFace~\cite{yi2014learning} and the refined version of MS-Celeb-1M~\cite{guo2016ms} provided by ~\cite{deng2019arcface}.
For the testing datasets,
we use the following benchmark datasets: Labeled Faces in the Wild (LFW)~\cite{huang2008labeled}, CALFW~\cite{zheng2017cross}, YouTube Faces (YTF)~\cite{wolf2011face},  IJB-B~\cite{whitelam2017iarpa}, IJB-C~\cite{maze2018iarpa}, and MegaFace~\cite{kemelmacher2016megaface}.

\textbf{Experiments setting}
For pre-prepossessing, we follow the recent papers~\cite{deng2019arcface,liu2017sphereface,wang2018cosface,kim2020broadface,deng2020sub} to generate the normalised face crops ($112\times112$).
For the backbone network,
we utilize the widely used neural networks(e.g.,ResNet-18, ResNet-50~\cite{he2016deep}),
in which we follow~\cite{deng2019arcface} to leverage BN-Dropout-FC-BN network structure to produce 256-dim embedding feature representation.
The size of the anchor image set and support set are empirically set as 50000 and 10, respectively.
We utilize the SGD algorithm with a momentum of 0.9 and weight decay of $5\times10^{-4}$.
For all the experiments,
we first pre-train the backbone network using the existing loss functions (\textit{e.g.}, ArcFace, CosFace, etc).
For the pre-training on the CASIA-WebFace,
the initial learning rate is 0.1,
and is divided by 10 at the 20k, 30k, 35k iterations.
The total iteration is 40k.
For the pre-training on MS-Celeb-1M,
the initial learning rate is 0.1 and divided by 10 at the 100k, 140k, 160k iterations.
The total iteration is 200k.
As for learning-based method in SSE,
we utilize two fully-connected with ReLU~\cite{nair2010rectified} layers as the regression network. The initial learning rate for the regression network is 0.001,and divided by 0.1 at the 15k,20k,25k iterations.
The total iteration is 30k. The batch size of all the experiments is set as 512. Besides, we use StyleGAN~\cite{karras2019style} to generate the anchor image set. In addition,
we use our proposed reference-based method and learning-based method in the Support Set Estimation module based on the pre-trained network,
which are called as \textbf{IDA-R} and \textbf{IDA-L}, respectively.
\subsection{Results on IJB-B and IJB-C datasets}
\vspace{-10pt}
\begin{table}[!htp]
\small
\centering
\caption{The results on IJB-B and IJB-C datasets when using different loss functions.}
\begin{tabular}{c|c|c|c|c}
\toprule
               \multirow{2}{*}{Method}&
               \multicolumn{2}{c|}{IJB-B (TAR@FAR)}&
               \multicolumn{2}{c}{IJB-C (TAR@FAR)}\\
               \cline{2-5}   
               & 0.001\% &0.01\% &0.0001\% &0.001\%\\
               \hline
               L2-Face~\cite{ranjan2017l2} &74.94&86.86&70.33&81.96\\
               +IDA-R &\textbf{76.89}&\textbf{87.45}&\textbf{73.87}&\textbf{82.87}\\
               +IDA-L &76.71&87.32&73.51&82.39\\
               \hline
               CosFace~\cite{wang2018cosface} &85.15&92.24&82.43&90.49\\
               +IDA-R &\textbf{86.46}&\textbf{92.59}&\textbf{83.45}&\textbf{90.61}\\
               +IDA-L &86.35&92.53&83.37&90.55\\
               \hline
               ArcFace~\cite{deng2019arcface} &85.50&93.09&80.61&90.62\\
               +IDA-R &\textbf{87.89}&\textbf{93.63}&\textbf{83.14}&\textbf{91.40}\\
               +IDA-L &87.78&93.54&82.90&91.38\\
               \bottomrule
 \end{tabular}
 \label{tab1}
\end{table}
This section evaluates our  newly proposed method on challenging IJB-B~\cite{whitelam2017iarpa} and IJB-C~\cite{whitelam2017iarpa} datasets.
Since our method can be readily integrated into different existing loss functions,
we provide detailed experiments based on L2-Face~\cite{ranjan2017l2}, CosFace~\cite{wang2018cosface} and ArcFace~\cite{deng2019arcface} using ResNet-50 based on MS-Celeb-1M~\cite{guo2016ms}.
As shown in Table~\ref{tab1},
we can observe that IDA-R and IDA-L methods both achieve significant performance improvement on IJB-B and IJB-C datasets in all cases when compared with original methods.
It indicates our proposed methods are robust for different loss functions.
Besides, the performance of our proposed IDA-L is comparable with IDA-R,
which means IDA-L can implicitly generate the local context information well.
\subsection{Results on LFW, CALFW and YTF datasets}
In this section,
to further demonstrate the effectiveness of our method,
we provide the results on LFW~\cite{huang2008labeled}, CALFW~\cite{zheng2017cross} and YTF~\cite{wolf2011face} in Table~\ref{tab2}.
Specifically,
we leverage our proposed IDA-R and IDA-L methods based on pre-trained ResNet-18 on  CASIA-WebFace dataset using ArcFace loss function~\cite{deng2019arcface}.
From Table~\ref{tab2},
when compared with the original method based on ArcFace,
our proposed IDA-R improves the accuracy by +0.30\% on LFW, +0.38\% on CALFW, and +0.26\% on YTF, respectively.
\vspace{-5pt}
\begin{table}[!htp]
    \centering
    \small
    \caption{
    The verification accuracy (\%) on the LFW, CALFW and YTF
datasets.
    }
    \vspace{-2pt}
    \label{tab2}
    \begin{tabular}{c|ccc}
        \toprule
        Methods  & LFW(\%)&CALFW &YTF(\%)  \\ 
        \hline
        ArcFace~\cite{deng2019arcface} &98.73&91.67&94.97\\
        +IDA-R &\textbf{99.03}&\textbf{92.05}&\textbf{95.23}\\     
        +IDA-L &98.98&92.03&95.17\\
            \bottomrule
    \end{tabular}
    \vspace{-5mm}
\end{table}
\subsection{Results on Megaface dataset}
In this section,
we evaluate the performance of our method on MegaFace~\cite{kemelmacher2016megaface},
which is a challenge benchmark for face identification.
We apply our newly proposed IDA-R and IDA-L methods on ResNet-100 model trained on MS-Celeb-1M dataset.
As shown in Table~\ref{tab1},
our proposed strategy also shows superiority on large-scale MegaFace face identification task.
Specifically,
when the size of distractor is $10^{6}$,
our proposed IDA-R method is better than  the ArcFace baseline by $0.21\%$.
Meanwhile,
our proposed IDA-L method also achieves similar performance with the competitive IDA-R method. 

\vspace{-5pt}
\begin{table}[!htp]
\centering
\small
\caption{Recognition accuracy on MegaFace under different sizes of MegaFace distractor.
}
\vspace{-2pt}
\begin{tabular}{c|c|c|c|c}
\toprule
               \multirow{2}{*}{Method}& \multicolumn{4}{c}{Size of MegaFace Distractor}\\
               \cline{2-5}   
                &$10^{3}$ &$10^{4}$&$10^{5}$&$10^{6}$\\\hline
                ArcFace~\cite{deng2019arcface} &99.62&99.37&98.95&98.30\\
                +IDA-R& \textbf{99.66}&\textbf{99.46}&\textbf{99.15}&\textbf{98.56}\\
                +IDA-L &99.66&99.45&99.13&98.54\\
               \bottomrule
 \end{tabular}
 \label{tab3}
 \vspace{-5mm}
\end{table}


\subsection{Ablation study}

	

\begin{figure}[!htb]
\begin{center}
	\subfloat[]{
		\label{fig:ab:a}
		\includegraphics[width=0.45\linewidth]{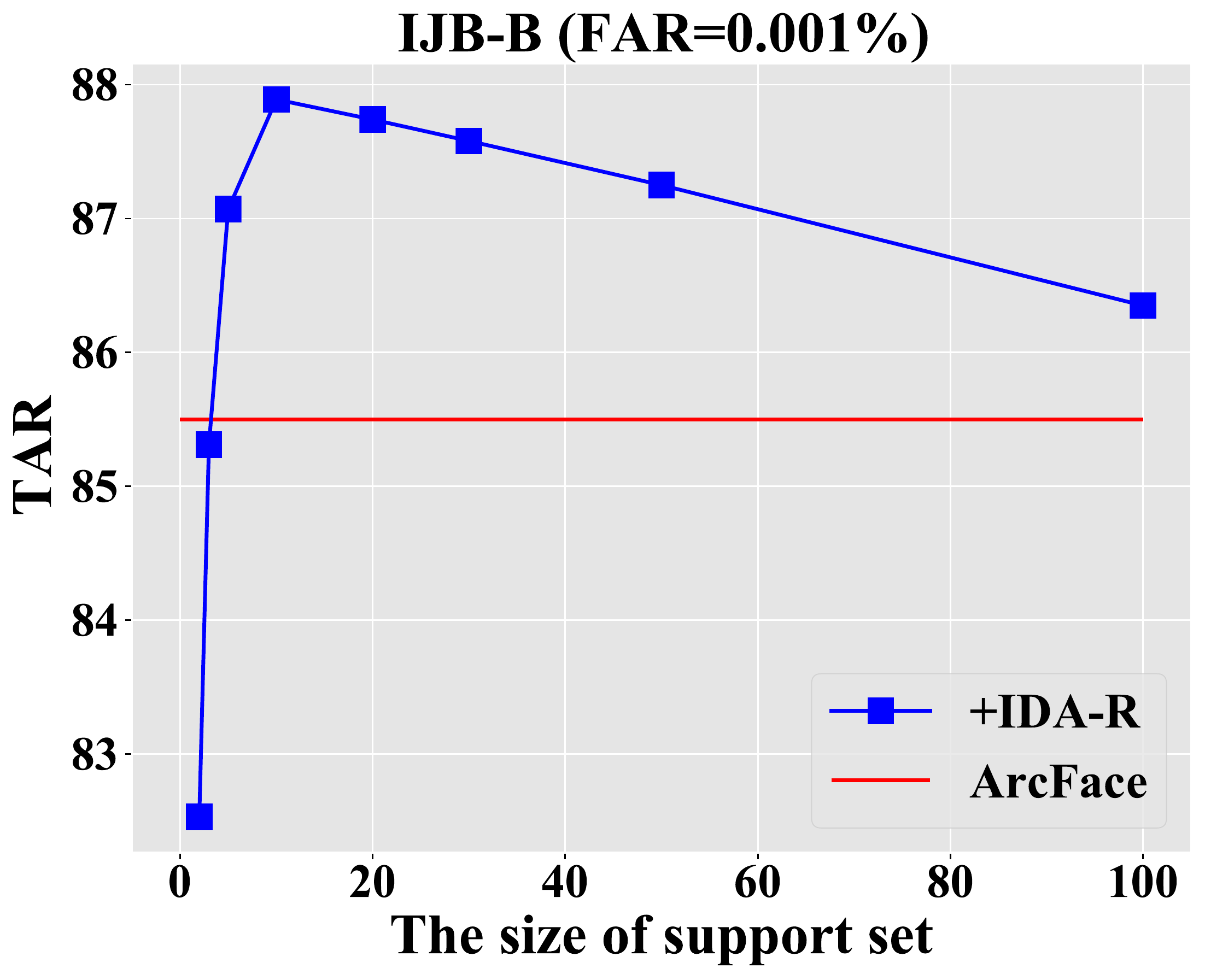}
	}
	\subfloat[]{
		\label{fig:ab:b}
		\includegraphics[width=0.45\linewidth]{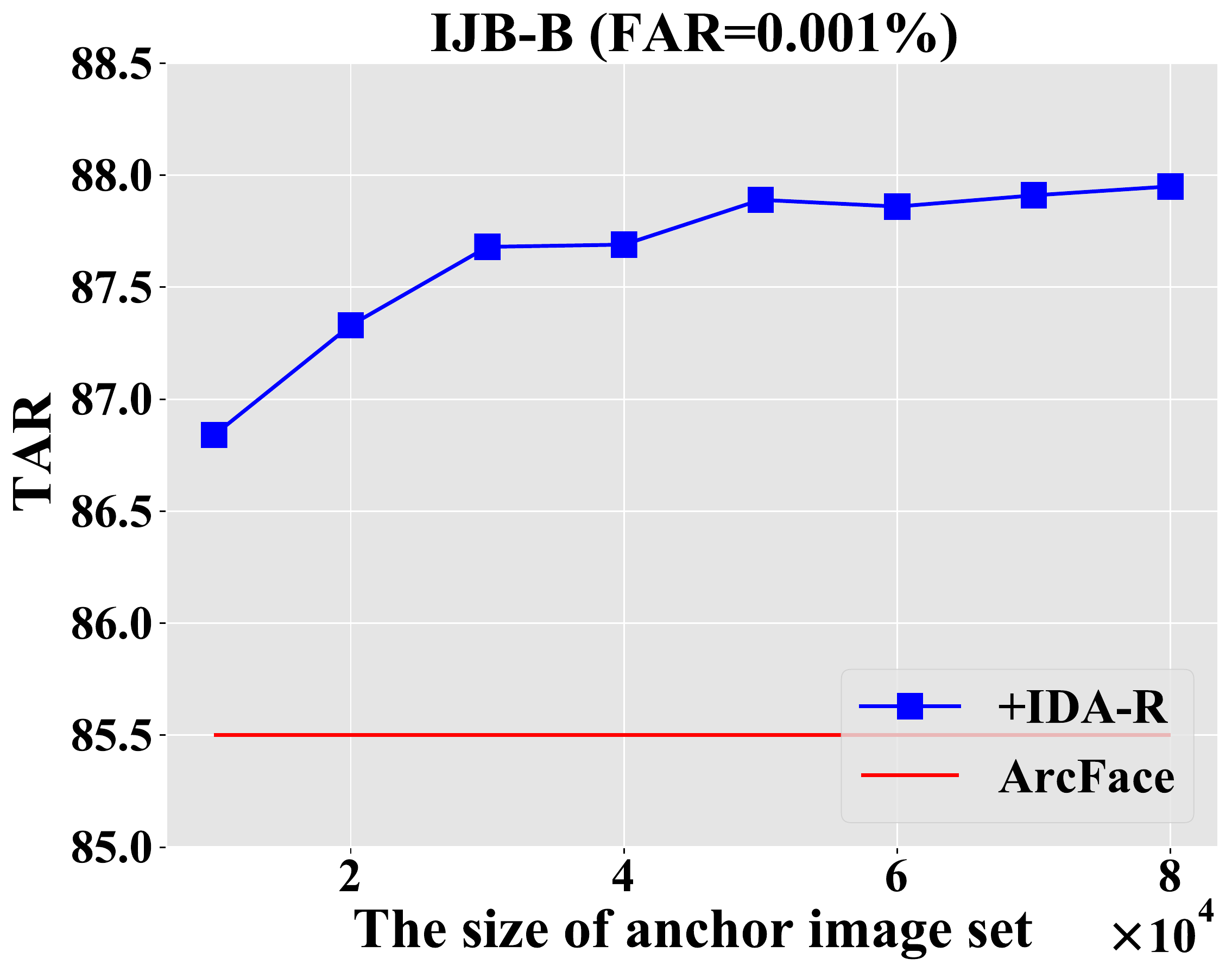}
	}
	\vspace{-10pt}
	\end{center}
	\vspace{-10pt}
\caption{\protect\subref{fig:ab:a} The effect of the size of support set.
\protect\subref{fig:ab:b} The effect of size of the anchor image set.}
\label{fig:ab}
\end{figure}

\noindent \textbf{The size of the support set and anchor embedding set.}
We evaluate our IDA-R method  using different sizes of support set and anchor embedding set, and results on the IJB-B dataset are shown in Fig~\ref{fig:ab}.
Specifically,
we leverage the ResNet-50 trained on MS-Celeb-1M dataset based on ArcFace loss function.
In Fig.~\ref{fig:ab:a},
we set the size of the anchor embedding set as 50000, and use different size of support set.
When the size of the support set increases from 1 to 10,
our method achieves the best performance,
which means that the support set could represent the inter-class local context appropriately.
However,
when the size of the support set continues to increase from 10,
we observe that the performance on the IJB-B dataset begins to drop. It is reasonable that as the size increases, the support set of each sample tends to be similar, and the local information is missing. 
Meanwhile,
in Fig.~\ref{fig:ab:b},
we set the size of anchor set as 10,
and use different sizes of anchor embedding set.
As the size of the anchor embedding set increases, the performance first gradually improves, then tends to be flat. The anchor embedding set is a sampling of the FR model's distribution in feature space. Increasing the size can obtain more accurate sampling and then get a better estimation of the support set. When the size is large enough, this improvement of performance gradually decreases.

\begin{figure}[t]
        \begin{center}
        \includegraphics[width=0.7\linewidth]{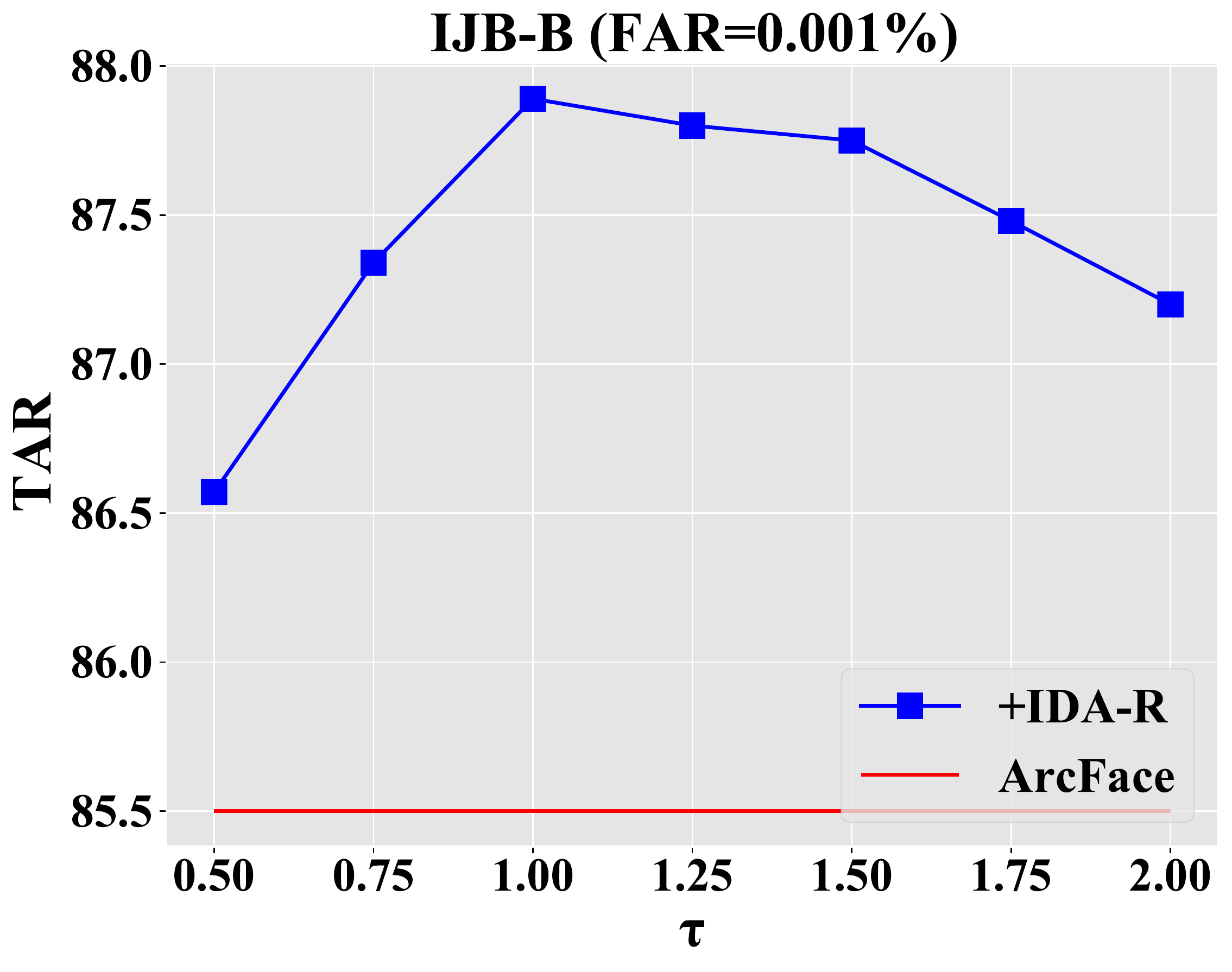}
         \caption{The effect of the hyperparameter $\tau$.
        }
        \label{tau}
        \end{center}
\end{figure}

\noindent \textbf{The effect of the hyperparameter $\tau$.}
We set $\tau=1$ in Eq.~\ref{eq:dao}.
To demonstrate the effect of the hyperparameter $\tau$,
we conduct more experiments by setting different values of $\tau$ on IJB-B dataset,
and the results at FAR=$0.001\%$ are shown in Fig.~\ref{tau}.
Specifically,
we leverage the ResNet-50 trained on MS-Celeb-1M dataset based on ArcFace loss function.
In Fig.~\ref{tau},
when the $\tau$ increases from 0.5 to 1,
the performance on IJB-B becomes better.
However,
when we continue to increase the $\tau$ from 1,
the performance begins to drop.
For this phenomenon,
we analyse that
in Eq.~4 of the main paper,
the $\tau$ controls the smoothness degree of the inter-class discrepancy for each image.
With $\tau \rightarrow 0$,
the inter-class discrepancy becomes indiscriminative.
While $\tau$ is increasing large,
the inter-class discrepancy is dominated by only a few restricted neighbor classes,
which cannot represent the local context information well.

\noindent \textbf{Different types of anchor embedding set.}
To analyze the effectiveness of the anchor embedding set,
we leverage different types of embedding for our IDA-R method,
and the results on IJB-B dataset are reported in Table~\ref{tab4}.
The ResNet-50 model is adopted and trained on MS-Celeb-1M dataset.
Specifically,
``ArcFace'' denotes the original result based on Arcface loss.
``Weights of FC layers'' denotes that we use the converged weight of the last fully-connected layer of ArcFace loss function. The weights bear conceptual similarities with the centers of
each face class.
``Real-db'' means that we randomly sample one image per identity from MS-Celeb-1M dataset, and extract features by the ResNet-50 model.
``Fake-db'' means that the images are generated by StyleGAN~\cite{karras2019style}, which trained on the MS-Celeb-1M dataset.
 
As shown in  Table~\ref{tab4}, similar results are achieved when using different types of anchor embedding set, which means that our method is not sensitive to the types of embeddings.
Furthermore,  the effectiveness of fake-db also shows that the specific ID is not important for estimating support set.
\begin{table}[H]
\centering
\small
\caption{The results on IJB-B dataset when using different types of anchor embedding set.}
\begin{tabular}{c|c|c}
\toprule
               \multirow{2}{*}{Types of anchor image set}& \multicolumn{2}{c}{IJB-B (TAR@FAR)}\\
               \cline{2-3}   
                &0.001\% &0.01\%\\\hline
                ArcFace~\cite{deng2019arcface} &85.50&93.09\\
                Weights of FC layers &87.50&93.48\\
                Real-db &87.86&\textbf{93.64}\\
                Fake-db &\textbf{87.89}&93.63\\
               \bottomrule
 \end{tabular}
 \label{tab4}
\end{table}
\vspace{-12pt}
\begin{figure}[!htp]
    \centering
    \includegraphics[width=1.0\linewidth]{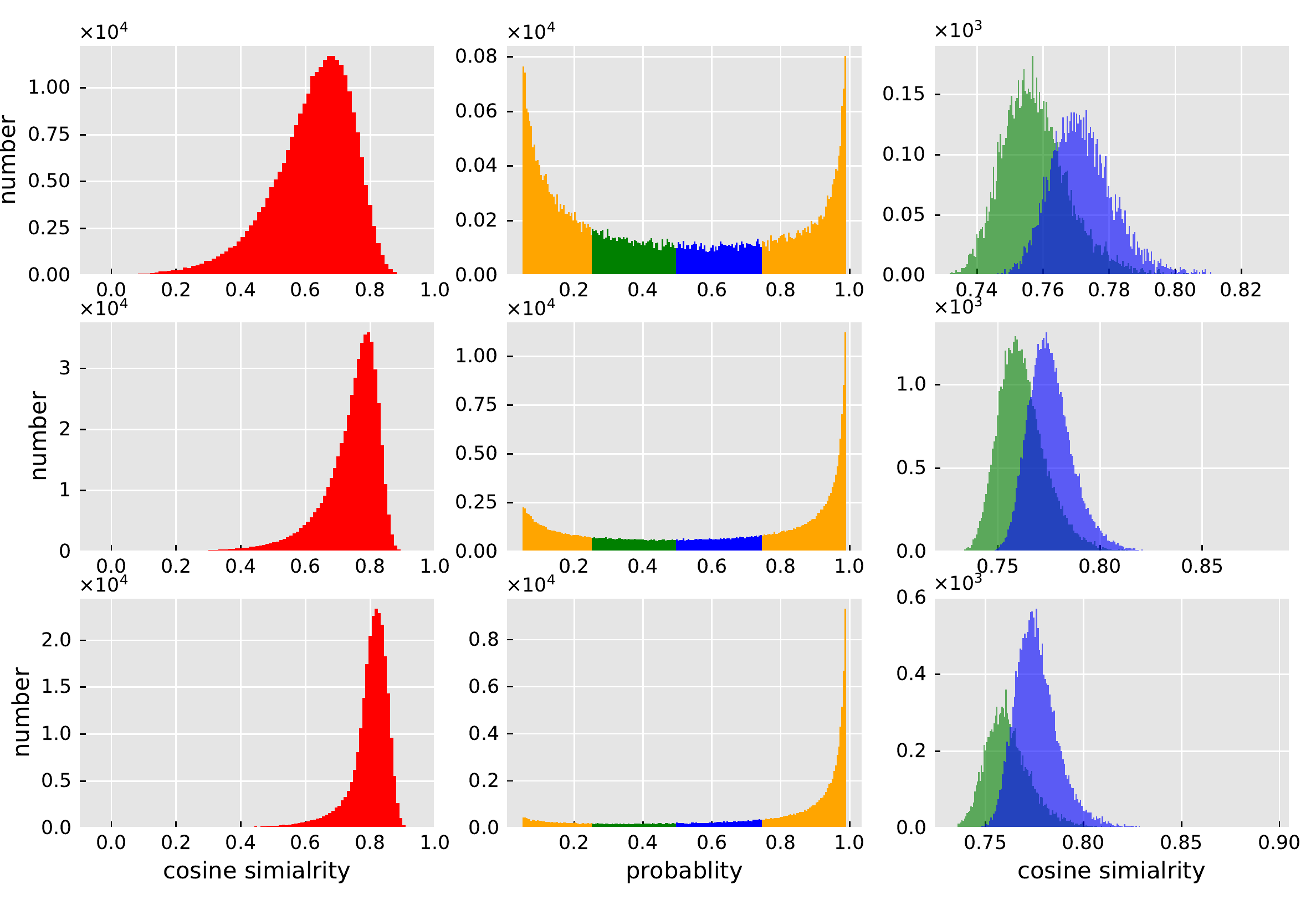}
     \caption{The distributions of probability (output of softmax) and cosine similarity  between the training sample and its positive category at different optimization stages. The entire training lasted 20,000 iterations, and the first row to the third row showed the distributions of the 80,000th, 140,000th, and 200,000th step. Two disjoint segments of the probability distribution are selected, and their corresponding cosine distributions are demonstrated in the last column.}
    \label{fig:distribution}
\end{figure}

\begin{figure}[!htp]
    \centering
    \includegraphics[width=1.0\linewidth]{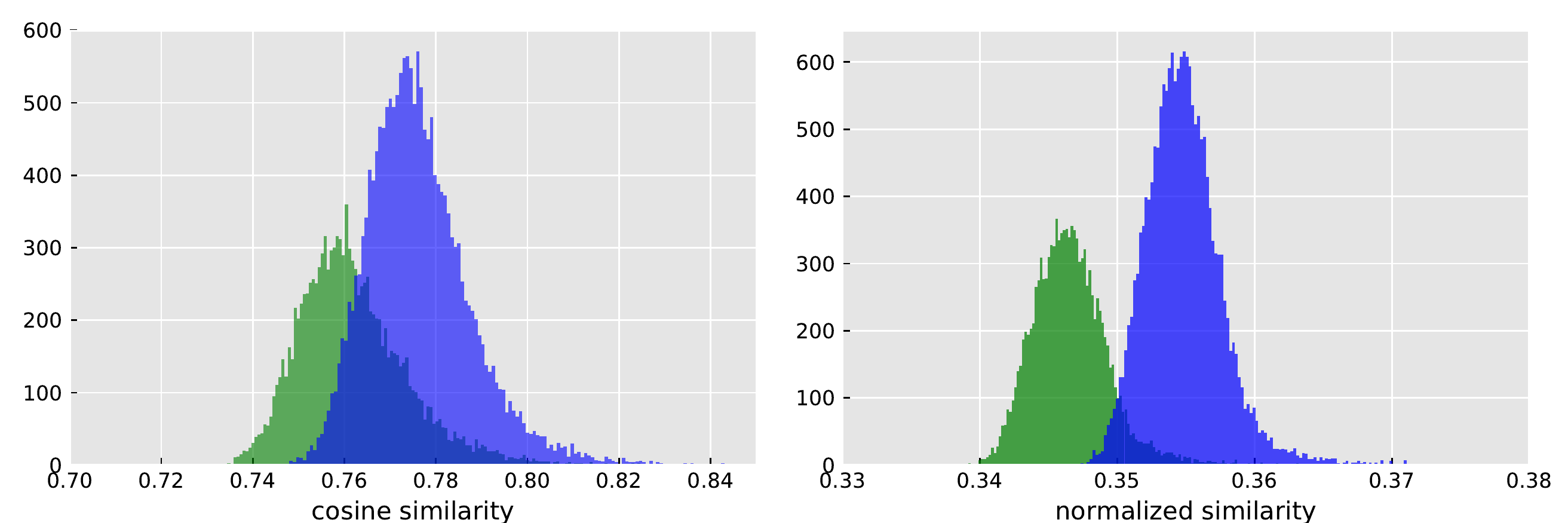}
     \caption{Comparison between cosine similarity and the normalized score outputted by IDA. The green and blue curves represent similarity distributions correspond to the two disjoint segments of probability, as described in Fig.~\ref{fig:distribution}. The right illustrates the original cosine similarity, and the left represents the normalized similarity.
    }
     \label{fig8}
 \end{figure}

\subsection{Further analysis} 
\noindent \textbf{Statistical analysis of cosine similarity and probability.} 
We train the ResNet-50~\cite{he2016deep} with MS-Celeb-1M~\cite{guo2016ms}.
As shown in the first two columns of Fig.~\ref{fig:distribution}, 
we visualize the 
cosine similarity score distribution corresponding to the positive category center and probability (output of softmax) distribution at different optimization steps. Then we select two disjoint segments in the probability distribution (the second column), and depict their corresponding cosine distributions in the last column of Fig.~\ref{fig:distribution}. 
The two cosine distributions overlap each other. 
We have two observations from Fig.~\ref{fig:distribution}:
1) As the training proceeds, the curves of both score distribution and probability become sharp. 
2) Meanwhile, the overlap of the selected distribution does not disappear with the convergence of the model, which shows the gap between cosine similarity and probability. 

\begin{figure}[!htp]
    \centering
    \includegraphics[width=1.0\linewidth]{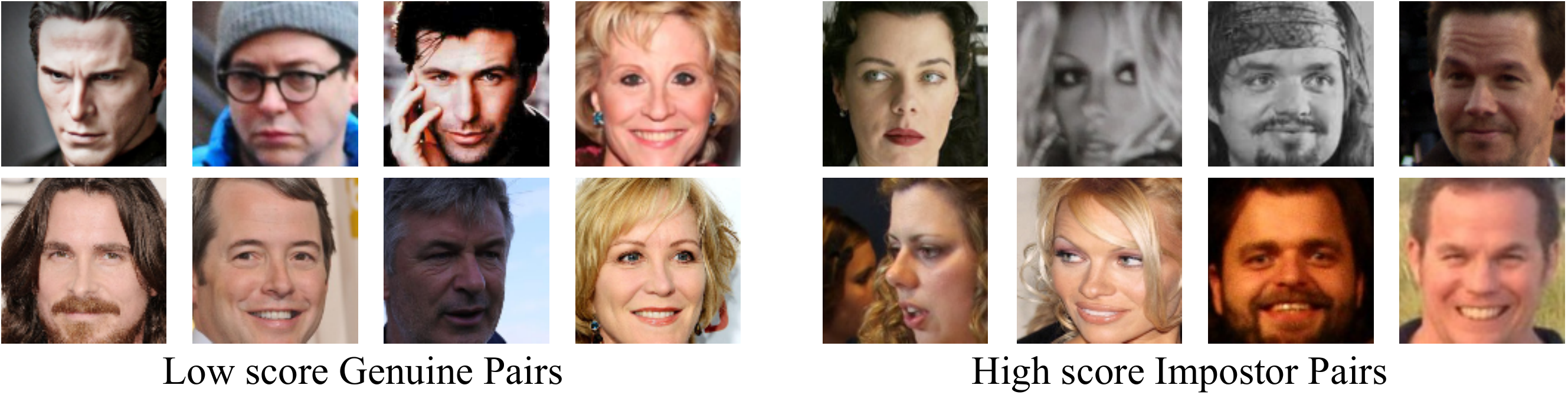}
     \caption{
        Two types of the false recognition samples from MegaFace dataset based on original cosine similarity, which are corrected by the proposed IDA.
    }
     \label{fig9}
 \end{figure}

\noindent \textbf{Effectiveness of IDA.}
To analyze the effect of the IDA, 
we re-visualize the normalized similarity (returned by DAO) 
of the samples in the two disjoint segments as selected above.
Compared with the P2P cosine similarity, the overlap of normalized score decreases, and the curve becomes sharper, as shown in Fig.~\ref{fig8}. It demonstrates that the misalignment between probability and similarity is reduced with the incorporation of inter-class discrepancy.
Fig.~\ref{fig9} illustrates some hard instances (wrongly labelled by conventional P2P way) corrected by IDA. 
\begin{table}[H]
\centering
\small
\caption{The results of different alternative methods on IJB-B and IJB-C datasets.}
\begin{tabular}{c|c|c|c|c}
\toprule
               \multirow{2}{*}{Methods}&
               \multicolumn{2}{c|}{IJB-B (TAR@FAR)}&
                \multicolumn{2}{c}{IJB-C (TAR@FAR)}\\
               \cline{2-5}   
                &0.001\% &0.01\% &0.0001\% &0.001\%\\\hline
                ArcFace~\cite{deng2019arcface} &85.50&93.09&80.61&90.62\\
                +PFE &86.63&93.32&82.79&91.11\\
                +IDA-R& 87.89&93.63&83.14&91.40\\
                +PFE\&IDA-R &\textbf{88.12}&\textbf{93.73}&\textbf{83.40}&\textbf{91.46}\\
               \bottomrule
 \end{tabular}
 \label{tab5}
\end{table}
\subsection{Integration with PFE}
As we know, existing work PFE~\cite{shi2019probabilistic} has explored the intra-class local context by representing each face image as distribution   and our proposed method aims to investigate the inter-class local context.
Therefore,
we can readily integrate our proposed reference-based method  IDA-R with PFE,
and the results on IJB-B and IJB-C datasets are as shown in Table~\ref{tab5}.
Specifically,
We reimplement the PFE based on ResNet-50 model trained by ArcFace loss function,
and apply our proposed method to the learned model by PFE.
From Table~\ref{tab5},
we observe that our proposed method IDA-R can produce better performance when integrated with the PFE method,
which further demonstrates the generalization ability of our proposed method.

\section{Conclusion}
In this paper,
we have investigated the inter-class local context information for further improvement of FR,
and propose a novel framework called IDA.
Extensive experiments among different FR benchmarks demonstrate the effectiveness of our proposed methods.

{\small
\bibliographystyle{ieee_fullname}
\bibliography{egbib}
}

\end{document}